\documentclass[sigconf, 10pt]{acmart}
\usepackage{balance}
\usepackage{graphicx}
\usepackage{subcaption}
\usepackage{xcolor}
\usepackage{multirow}
\usepackage{array}
\usepackage{breqn}
\usepackage{mathtools}
\usepackage{amsmath}
\usepackage[ruled,vlined,linesnumbered]{algorithm2e}
\usepackage{tikz}

\newcommand{\logicand}{\;\&\;}
\newcommand{\logicor}{\;|\;}
\labelformat{subfigure}{\thefigure(#1)}
\newcolumntype{P}[1]{>{\centering\arraybackslash}p{#1}}
\AtBeginDocument{%
  }

\setcopyright{none}
\begin{document}

\title{AutoLife: Automatic Life Journaling with Smartphones and LLMs}


\author[Huatao Xu, Panrong Tong, Mo Li, Mani Srivastava]{Huatao Xu$^{1}$, Panrong Tong$^{2}$, Mo Li$^{1}$, Mani Srivastava$^{3}$ \thanks{Mo Li is the corresponding author.}}
\affiliation{%
  \institution{$^{1}$Hong Kong University of Science and Technology, $^{2}$Alibaba Group, \\$^{3}$University of California Los Angeles 
  }
  \country{Email:huatao@ust.hk, panrong@alibaba.inc, mbs@ucla.edu, lim@cse.ust.hk}
}
\def \authors{Huatao Xu, Panrong Tong, Mo Li, and Mani Srivastava}


\begin{abstract}
This paper introduces a novel mobile sensing application - \textit{life journaling} - designed to generate semantic descriptions of users' daily lives. We present AutoLife, an automatic life journaling system based on commercial smartphones. AutoLife only inputs low-cost sensor data (without photos or audio) from smartphones and can automatically generate comprehensive life journals for users. To achieve this, we first derive time, motion, and location contexts from multimodal sensor data, and harness the zero-shot capabilities of Large Language Models (LLMs), enriched with commonsense knowledge about human lives, to interpret diverse contexts and generate life journals. To manage the task complexity and long sensing duration, a multilayer framework is proposed, which decomposes tasks and seamlessly integrates LLMs with other techniques for life journaling. This study establishes a real-life dataset as a benchmark and extensive experiment results demonstrate that AutoLife produces accurate and reliable life journals.

\end{abstract}
\keywords{Mobile Sensing, Life Journaling, Large Language Model}


\maketitle
\newcommand{\sysname}{AutoLife\xspace}
\section{Introduction}
\begin{figure}[t!]
  \centering
  \includegraphics[width=0.95\linewidth]{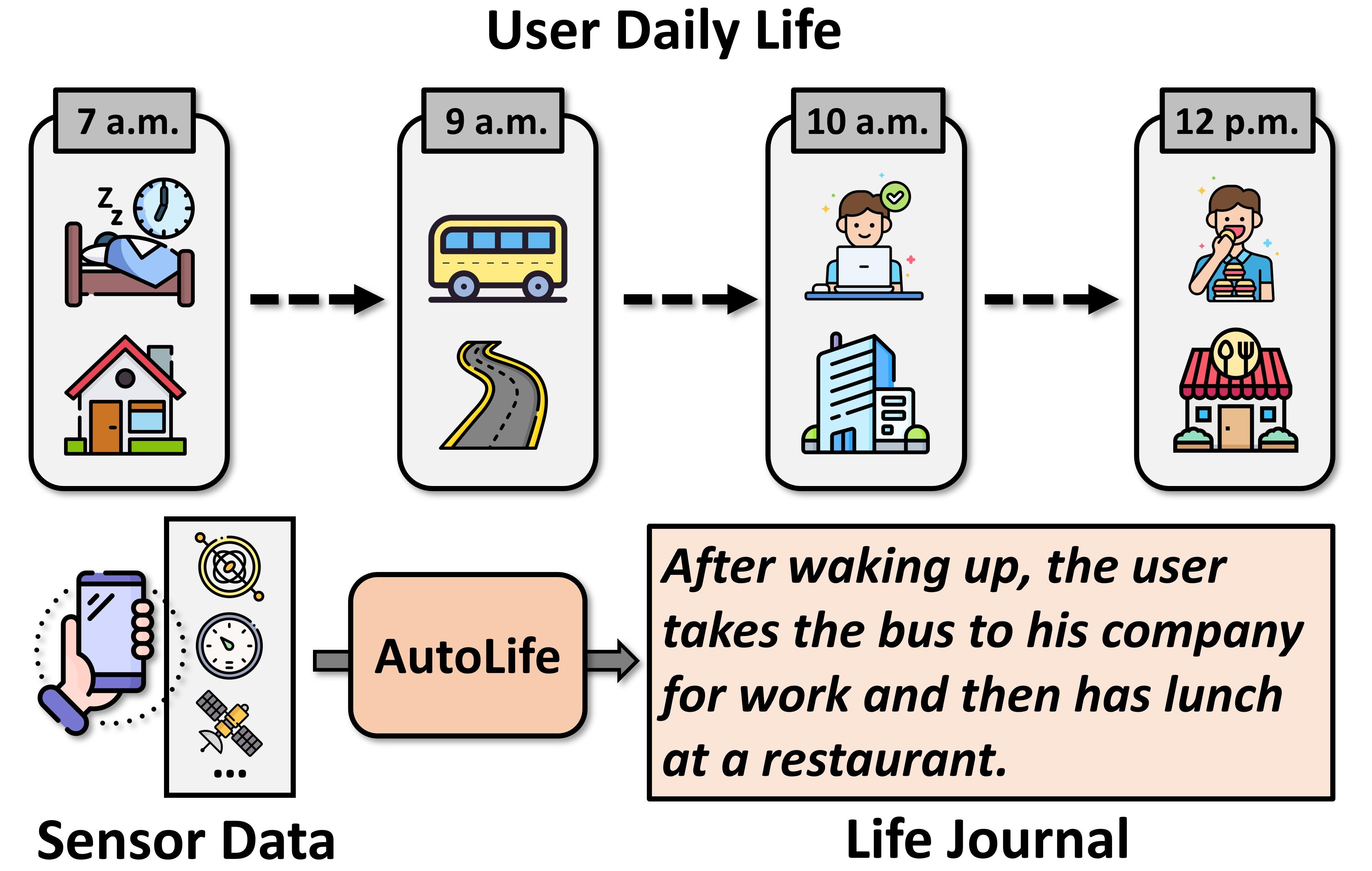}
  \caption{Life journaling application.}
  \label{fig:application}
  \vspace{-0.5cm}
\end{figure}

The widespread adoption of mobile devices like smartphones has significantly transformed many aspects of daily life. Beyond traditional mobile applications, this paper introduces a novel mobile sensing application named "\textbf{Life Journaling}" -- \textit{an approach to automatically generate detailed semantic descriptions of a person's daily life}. Figure \ref{fig:application} presents an example of a journal generated from such an envisioned life journaling application, which offers natural and semantic descriptions of the person's life context including key activities, behaviors, and circumstances in a comprehensive way. We believe life journaling is a very useful application and can support numerous downstream use cases, including personalized recommendations based on user behaviors, automatic annotation or organization of personal photos or video clips based on daily lives, optimizing daily routines for health, and many more.

Unfortunately, to the best of our knowledge, there is no existing solution for such a valuable application at present. Existing lifelogging systems \cite{wikipedia_lifelog, gurrin2014lifelogging, ksibi2021overview} focus on recording daily life as raw digital data such as videos or sensor readings rather than understanding high-level life semantics. Prior human activity recognition (HAR) studies \cite{vrigkas2015review, zhang2017review, xu2021limu,zhou2020xhar,hong2024crosshar, xu2023practically} attempt to identify user activities by predicting motion labels like "walking" or "jogging", which are far less informative compared to generating rich life contexts as targeted by life journaling. While there are several commercial digital journaling apps, such as Day One \cite{dayone_journal_app} and Journal \cite{apple_2023_journal}, they are not designed to automatically generate journals and rely heavily on human inputs. So, there is a significant gap in building a viable life journaling system at present.

To fill the gap, this paper presents \textbf{\sysname}, an automatic life journaling system that generates journals of users' daily lives based on smartphone sensor data. A key feature is that \sysname requires no user input — all a user needs to do is to carry their own smartphone while going about their activities. As shown in Figure \ref{fig:application}, \sysname processes various sensor readings and other data sources (without photos or audio) accessible from the smartphone, outputting detailed journals of the user's daily life. An essential challenge faced in developing such a system is \textbf{\textit{how to fuse those multimodal sensor inputs and generate accurate yet open-vocabulary semantic descriptions?}}

To the best of our knowledge, there is no existing dataset for this specific task, making conventional deep-learning solutions inapplicable. Extensive human life knowledge may be required to interpret diverse contexts, e.g., motion and time, and accordingly infer complex human behaviors. This paper builds on our key observation that such context interpretation and inference tasks align well with the strengths of Large Language Models (LLMs), which are trained on large-scale text corpora and possess extensive commonsense knowledge of human behaviors. However, directly using LLMs to analyze sensor data for life journaling can result in hallucinations or low-quality journals due to the high complexity of the task. To address this, our key approach is to extract rich and accurate contexts from sensors, fuse them as flexible texts, and leverage LLMs to synthesize comprehensive life journals from these contextual inputs. Two technical challenges are addressed in the design of \sysname.


First, we must address a critical question, namely, \textit{what information is desired to derive accurate life journals and how such information can be extracted from various data sources}? While numerous HAR studies \cite{gong2019metasense, xu2021limu, jain2022collossl, xu2023practically,hong2024crosshar} have been conducted, we notice that they typically produce only basic motion labels, such as "stationary" or "walking", due to limitations in sensor datasets and the constraints of motion sensors. Such motion contexts can provide some insights into user behaviors but are insufficient for generating a comprehensive journal. In \sysname, we incorporate two additional contexts - time and location. Both are instrumental in understanding user behaviors, as illustrated in Figure \ref{fig:application}. For instance, if a user remains stationary at a restaurant during mid-noon, it can be reasonably inferred that they are likely having lunch. To detect location context, we exploit GPS locations with geographic information systems (GIS), e.g., the Google Maps Platform \cite{GoogleMaps}. While existing APIs do not reveal comprehensive location contexts, in \sysname we propose to utilize large vision-language models (VLMs) like GPT-4o \cite{OpenAI2024GPT4o} to generate location context by interpreting map segments queried from GIS. We also incorporate WiFi SSID information and leverage lighter-weight LLMs like GPT-3.5 to further infer the user's surrounding environment (often when indoors).

Second, to further improve the quality of journals, we build special enhancements around the LLMs, including providing journal examples in the prompts and utilizing two LLM-based modules to pre-process the contexts and post-process the generated journals. Specifically, we address a key challenge of \textit{how to assist LLMs in handling lengthy sensor data collected over long daily life periods?} Different from existing HAR applications interested in labeling short periods of activities \cite{banos2014window} like a few seconds, life journaling typically spans a much longer duration over hours, which adds not only complexity to the task, but also difficulties to LLMs in handling the lengthy inputs. To address this challenge, we design a multi-layer framework that breaks life journaling into smaller and manageable subtasks. \sysname first segments the sensor data into small windows and extracts both motion and location contexts from these segments with the combined use of conventional signal processing or LLM/VLMs. In the middle layer, \sysname represents the contexts as text, which are then fused and refined before being sent for comprehension by the LLMs. In the last layer, the refined contexts with reduced lengths are consolidated, encapsulating extended-duration context, and finally fed to LLMs to generate the final journals. A duty-cycled data collection approach is applied to further reduce system overhead.

The proposed \sysname system is prototyped and evaluated with a self-collected human life dataset that contains diverse behaviors like hiking, cycling, shopping, and working of 3 volunteers in Hong Kong. An Android app is developed to continuously collect sensor data from smartphones while users go about their daily activities. For each experiment, the volunteer manually creates reference journals, consisting of text descriptions of the volunteer's behaviors. To evaluate the qualities of journals generated by \sysname, we compare the similarities between them with the reference journals using metrics such as BERTScore \cite{zhang2019bertscore}. Our extensive experiments demonstrate that some LLMs like Claude 3 with our system can achieve an average BERTScore F1 higher than 0.7. In summary, this paper makes the following contributions:
\begin{enumerate}
    \item The paper for the first time showcases a novel mobile sensing application that can automatically generate life journals with commercial smartphones.
    \item We present the first life journaling system, \sysname,  which creatively incorporates both LLM/VLMs and conventional signal processing to fuse various sensor data and synthesize long-duration life journals.
    \item The system is prototyped and comprehensively evaluated. The dataset we establish will be made publicly available and may serve as a benchmark for future research on this topic.
\end{enumerate}
The rest of this paper is organized as follows. Section 2 presents the related works. Sections 3-6 introduce the design of \sysname. Section 7 provides implementation and evaluation results. Section 8 discusses and Section 9 concludes this paper.

\section{Related Works}
\subsection{Life Logging}
Lifelogging \cite{wikipedia_lifelog, gurrin2014lifelogging, ksibi2021overview,gemmell2006mylifebits} is a technique that digitizes human daily life, which can support many applications, including health monitoring and memory enhancement. With the rapid proliferation of mobile devices, many mobile devices or applications have been developed for lifelogging. For example, Microsoft's SenseCam \cite{microsoft_sensecam} is a pioneering wearable camera designed to capture continuous photographic or video records of a person’s day. However, most lifelogging works aim at `logging' the user's daily life instead of generating high semantic journals. Additionally, many solutions require wearable cameras \cite{zhang2017review, gurrin2016ntcir, bolanos2016toward} or smart glasses \cite{jiang2019memento}, which are not ubiquitous and introduce extra costs.
\begin{figure}[t!]
  \centering
  \includegraphics[width=0.95\linewidth]{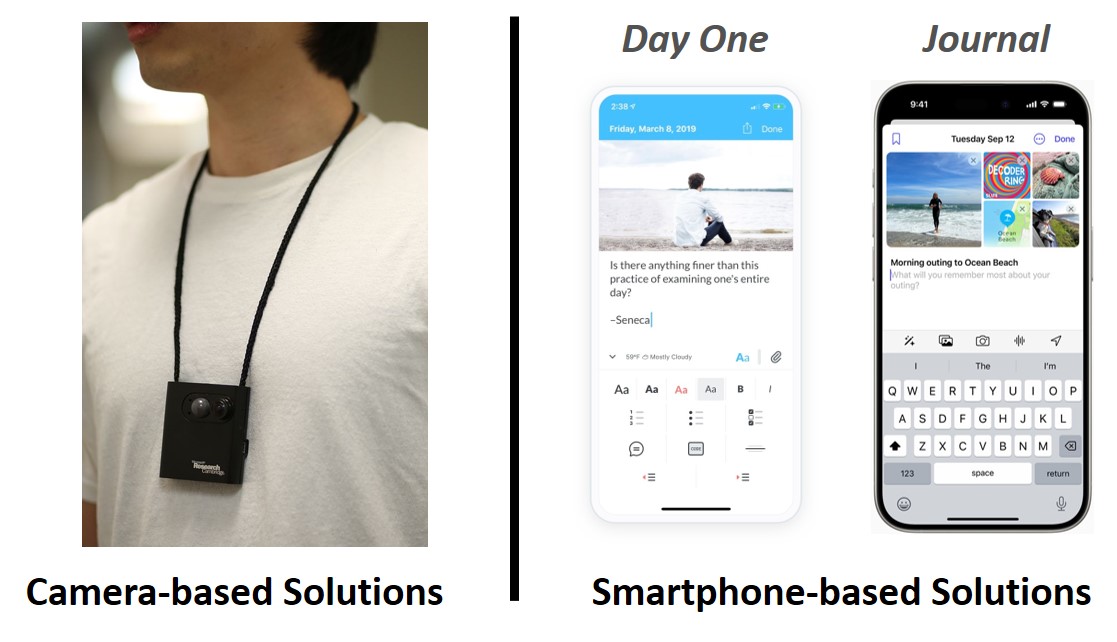}
  \caption{Existing lifelogging solutions. Left shows a user wears SenseCam \cite{glogger_2013} while right shows two digital diary applications, i.e., Day One \cite{dayone_journal_app} and Journal \cite{apple_2023_journal}.}
  \label{fig:lifelogging}
  \vspace{-0.5cm}
\end{figure}


Smartphones are widely available and there are numerous digital journaling applications on the market, as illustrated in Figure \ref{fig:lifelogging}. However, all these apps require extensive manual input from users. A recent work, MindScape \cite{nepal2024contextual} proposes to generate personalized prompts with LLMs, such as "Your running routine has really taken off! How’s that influencing your day?" and records the user's responses for journaling, which still requires user input. Unlike existing solutions, our approach generates life journals for users by leveraging data collected from ubiquitous devices like smartphones, eliminating the need for manual input.

\subsection{Activity Recognition}
Beyond lifelogging, Human activity recognition (HAR) is a critical research topic that aims at recognizing users' daily activities like `answering the phone' or `walking'. There are extensive HAR studies and wearable-based solutions \cite{jain2022collossl,hong2024crosshar,saeed2019multi,liu2020giobalfusion, zhou2020xhar, qin2019cross, xu2021limu, gong2019metasense,xu2023practically} can be implemented on off-the-shelf smart devices and are more ubiquitous compared with vision-based \cite{vrigkas2015review, zhang2017review, poppe2010survey} or wireless-based \cite{yu2019rfid, zhao2017learning, jiang2018towards} solutions. 


Despite significant progress in the field, several limitations persist: (1) Most existing methods \cite{gong2019metasense,xu2021limu,zhou2020xhar,hong2024crosshar, xu2023practically,jain2022collossl,ji2024hargpt} rely solely on motion sensors like inertial measurement units (IMUs), which are insufficient for distinguishing complex activities. For example, IMU data may only indicate that a user remains stationary for an extended period, without providing enough context to determine whether they are having a meal or attending a class. (2) No existing HAR models can generally and accurately recognize a wide range of motion types, primarily due to the lack of large-scale and comprehensive datasets. More importantly, motion labels obtained from existing HAR methods, such as `walking' or `cycling', do not provide the comprehensive information that life journals offer. In summary, current HAR approaches fall short of achieving the goals of life journaling.



\subsection{Context Awareness}
Location awareness refers to the ability of devices to detect their geographical positions while context awareness \cite{liu2011survey,yurur2014context} extends beyond simple geographical location, allowing devices or systems to interpret various aspects of their environment. Understanding location context is crucial for sensing user behaviors; for example, if a user remains stationary in a restaurant for an extended period, they are likely having a meal. In this paper, we explore a specific aspect of context awareness -- "detecting the location context of devices" such as identifying whether a device is at a restaurant or a park. One approach might involve leveraging computer vision models to analyze photos and derive location contexts or scenes \cite{tahmasebi2020machine,wang2015places205,zhou2014learning}. However, it is impractical to expect users to continuously capture photos to generate journals. Instead, this paper introduces a novel method to derive location contexts using low-cost and easily accessible sensor data from smartphones.

\subsection{LLM-based Sensing} 
Large Language Models (LLMs) have achieved remarkable advancements across a wide range of tasks \cite{brown2020language, scao2022bloom, zeng2022glm,openai2023gpt4,touvron2023llama,generalpatternmachines2023}. These out-of-the-box capabilities demonstrate that LLMs contain vast amounts of world knowledge, acquired through extensive training on large-scale text datasets. Some works \cite{driess2023palm,peng2023kosmos,jiang2022vima,brohan2023rt,ye2023mplug,wang2023visionllm, OpenAI2024GPT4o} extend LLMs into multimodal models, such as vision language models (VLMs) \cite{jiang2022vima}, to tackle various image-related tasks. Additionally, several studies introduce innovative LLM applications in diverse fields, such as Liu et al.'s work \cite{liu2023large}, which analyzes medical data for health-related tasks. Notably, researchers have proposed the concept of Penetrative AI \cite{xu2023penetrative}, exploring the integration of LLMs with the physical world through IoT sensors. With embedded extensive commonsense knowledge, LLMs/VLMs can perform physical tasks by analyzing IoT signals, such as detecting heartbeats using digitized or figure-based ECG data \cite{xu2023penetrative}. Inspired by the idea of Penetrative AI, we propose a new application of LLMs/VLMs for deriving life journals from sensor data on smartphones.
\section{\sysname}
\subsection{Problem Definition}
In this paper, we introduce a new application called \textbf{life journaling}, which generates journals for users' daily lives through mobile devices. We assume that our system functions as a mobile application on these devices, with regular access to sensor data. The system takes low-cost and long-term sensor data as input, such as accelerometer readings or GPS locations. The output is a series of sentences that accurately describe the user's daily activities, e.g., visiting a museum or resting at home.
\subsection{Overview}
\begin{figure}[t!]
  \centering
  \includegraphics[width=1.00\linewidth]{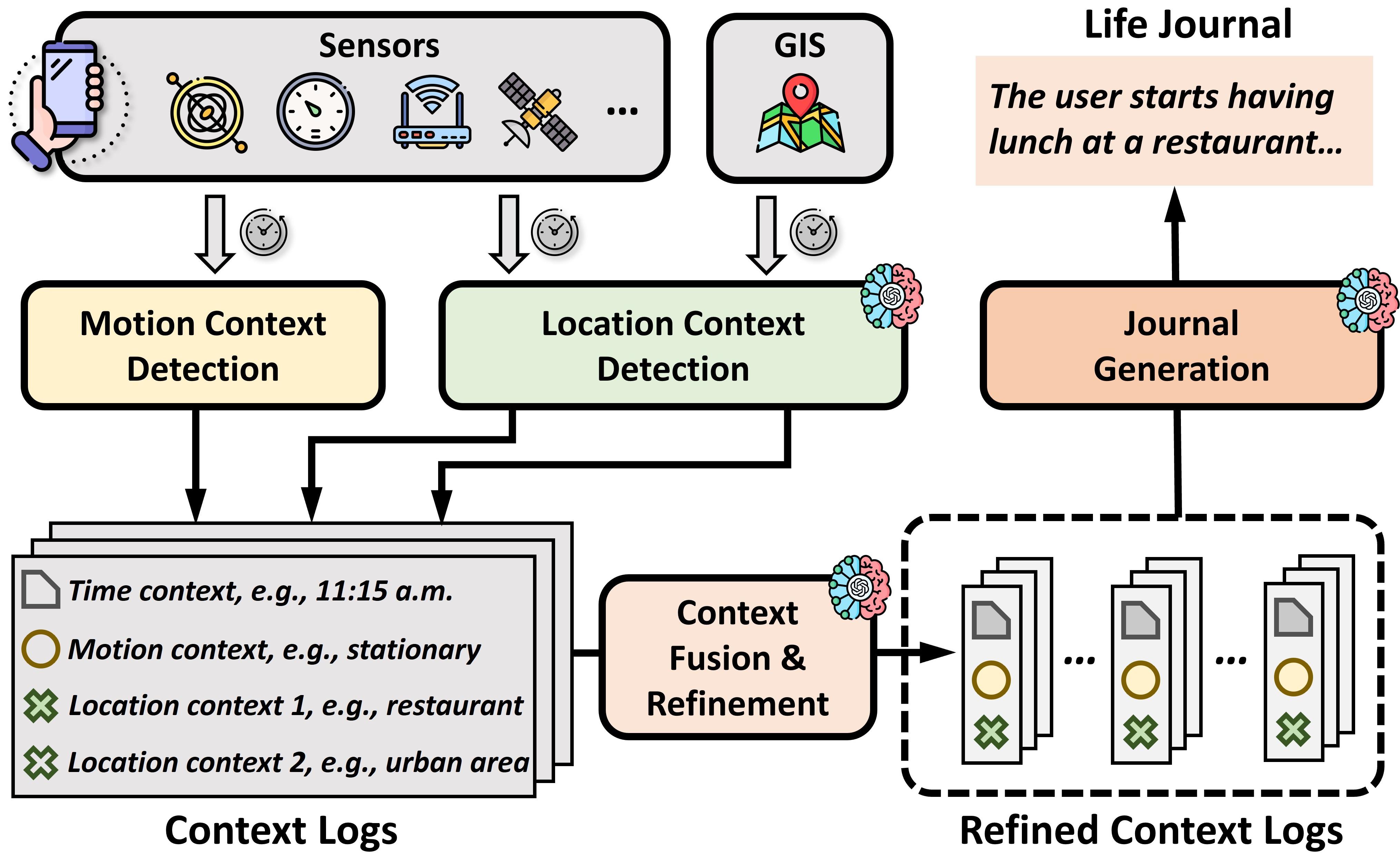}
  \caption{\sysname overview.}
  \label{fig:overview}
  \vspace{-0.5cm}
\end{figure}
Figure \ref{fig:overview} presents the overview of \sysname. Instead of directly feeding long-duration sensor data to LLMs for life journaling, that may cause hallucinations and low-quality journals, \sysname optimizes the use of LLMs with various sensor data by a multi-layer framework that decomposes the life journaling task process into manageable subtasks, each addressed by specialized modules. First, \sysname periodically accesses sensor data from smartphones in short periods. The \textit{motion context detection} and \textit{location context detection}, are designed to derive the user's contexts from multiple sensor resources. Particularly, \textit{location context detection} presents a novel approach to obtain accurate and general location contexts using LLMs or VLMs. Next, \sysname represents these contexts as flexible texts and utilizes another LLM-based module to enhance their precision and reduce text length. Finally, \sysname aggregates the enhanced context logs over a long duration and processes them through the \textit{journal generation} module, where LLMs synthesize the information to generate comprehensive life journals for users.
  
\subsection{Input Sensors}
It is intuitive that any single sensor data, e.g., the accelerometer or GPS location, cannot provide sufficient information to infer accurate journals. Therefore, our system integrates data from multiple sensors. Below is an overview of the chosen sensor features and how they are pre-processed.
\begin{itemize}
    \item \textbf{Accelerometer} sensors capture the device's accelerations. We use step-count algorithms \cite{android_step_counter_sensors} to estimate the user's steps from a duration of accelerometer readings, which serves as another important indicator.
    \item \textbf{Gyroscope} measures the device's angular velocity, which can be integrated with the accelerometer to estimate device orientation. The human-caused acceleration \cite{android_motion_sensors} is also an important feature, which can be computed by fusing the two sensors.
    \item \textbf{Barometer} measures air pressure, which can be used to estimate rough altitude using the barometric formula \cite{andrews2010introduction}. We then compute the altitude change over a time period as $\Delta h = h_i - h_j$, where $h_i$ represents the altitude at time $i$. The altitude change is a valuable feature for detecting user movement.
    \item \textbf{GPS speed} reflects the user's movement on the horizontal plane. Since satellite signals may be blocked when the user is indoors, the speed reported by the localization module can be unreliable. We filter GPS speed data when the number of detected satellites is fewer than 5.
    \item \textbf{GPS location} provides the geographic coordinates, consisting of latitude and longitude. Similarly, GPS data can be unreliable indoors and we filter out locations where the horizontal accuracy radius, as reported by the Android API \cite{android_location}, exceeds 50 meters. 
    \item \textbf{WiFi} signals can also help determine the user's location and are used for localization in the Google Fused Location Provider \cite{google_fused_location}. Recent studies \cite{ni2022experience, xu2023penetrative} have shown that WiFi Service Set Identifiers (SSIDs) can offer valuable insights into a user's surroundings.
\end{itemize}
Note that during implementation, we access the geographic location from Andoird Fused Location Provider API \cite{google_fused_location}, which fuses multiple sources including GPS and WiFi for more accurate localization.
\section{Context Detection}
This section will elaborate on how we fuse the input sensors and derive motion or location contexts for life journaling.

\subsection{Motion Context}
Motion information like walking is a key indicator for determining users' behaviors. Extensive research in HAR \cite{gong2019metasense,xu2021limu,zhou2020xhar,hong2024crosshar,xu2023practically,jain2022collossl,ji2024hargpt,malekzadeh2019mobile,xu2023penetrative} has demonstrated the potential of leveraging motion sensors to identify activities like jogging or cycling. However, these approaches cannot be directly applied to life journaling because most available public datasets \cite{stisen2015smart,malekzadeh2019mobile,shoaib2014fusion,zhang2012usc} cover only a limited range of sensor modalities, users, devices, and labeled data, making it challenging to build general models for recognizing activities.


To build a general solution, we propose a new rule-based motion detection algorithm by exploiting multimodal sensors. As outlined in Algorithm \ref{al:motion}, our approach fuses multiple features post-process by raw sensor data, including step counts, acceleration excluding gravity, altitude change, and GPS horizontal speed. The rules are based on commonsense knowledge; for example, if the step count is low while the speed is high, the user is likely using transportation. Despite leveraging multiple sources, ambiguities still arise when determining certain activities, so our algorithm acknowledges the limitations of sensors and can output multiple possible motions when the input data is inconclusive, e.g., `escalator/elevator'. Later, we leverage LLMs to reduce these ambiguities by incorporating location context.



\begin{algorithm}[t!] 
\SetKwInOut{Input}{Input}
\SetAlgoLined
\KwIn{step count $s$ per minute, acceleration excluding gravity $a\, \text{m/s}^2$, altitude change $\Delta h\, \text{m}$, horizontal speed $v\, \text{m/s}$.}
\KwOut{motion list $L$.}
$L \gets []$\;  
\uIf{$s <= 2\logicand a <= 0.1\logicand |\Delta h| <= 0.1\logicand v <= 0.1$ }{
    $L \gets L + [\text{`stationary'}]$\;
}
\ElseIf{$s <= 10\logicand |\Delta h| <= 1.0\logicand v < 0.5$ }{
    $L \gets L + [\text{`limited motion'}]$\;
}
\uIf{$s >= 140\logicand 2.0 <= v <= 5.0$ }{
    $L \gets L + [\text{`jogging/running'}]$\;
}
\uIf{$s >= 50\logicand v < 1.8$ }{
    $L \gets L + [\text{`walking'}]$\;
}
\uIf{$s >= 50\logicand v >= 4.0$ }{
    $L \gets L + [\text{`cycling'}]$\;
}
\uIf{$(s <= 5 \logicand v > 2) \logicor v > 5 $}{
    $L \gets L + [\text{`vehicle/subway/ferry/train'}]$\;
}
\If{$s <= 10 \logicand \Delta h > 2.5 \logicand v < 2$}{
    $L \gets L + [\text{`escalator/elevator'}]$\;
}
\Return $L$\;

\caption{Motion detection algorithm in \sysname.}
\label{al:motion}
\end{algorithm}

We reference thresholds in gait and activity analysis studies \cite{tudor2005pedometer,bohannon1997comfortable,vanini2016using} and fine-tune them using our dataset. We evaluate the proposed algorithm on our dataset (detailed in Section \ref{sec:imp}) and Sussex-Huawei Locomotion Dataset \cite{gjoreski2018university}, where certain labels like `car' and `bus' are merged. The algorithm is tested on 831 samples from our dataset and 13,544 samples from the Sussex-Huawei dataset. The results show that the algorithm achieves an average precision of 0.864 and 0.773 on the two datasets, respectively. Further evaluation experiments demonstrate that the motions recognized by our algorithm can enhance the quality of life journals.

\subsection{Location Context}
Location context is also crucial for accurately inferring a user's activity. However, detecting location contexts using ubiquitous sensors on smartphones is not straightforward. In this section, we design a low-cost solution for detecting location contexts.

\subsubsection{Location Context from GPS location}
\begin{figure*}[t!]
  \centering
  \includegraphics[width=\linewidth]{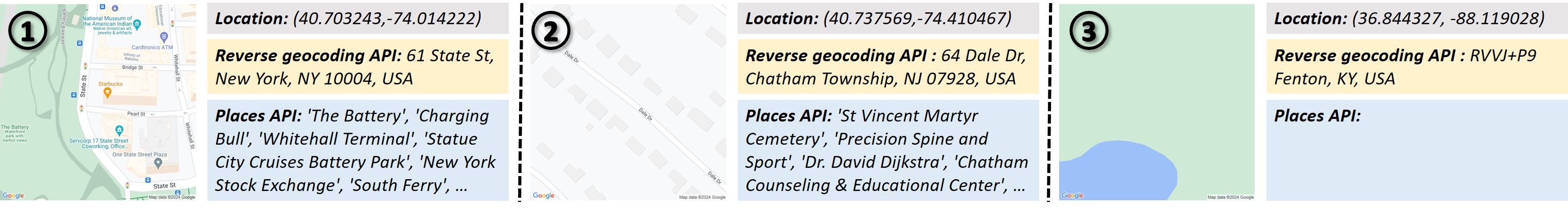}
  \caption{Examples of detecting location contexts with address and places. Results are from Google Maps Geocoding and Places API, respectively. The left side shows the map segments centered at corresponding locations.}
  \label{fig:location:other}
\end{figure*}
\begin{figure*}[t!]
  \centering
  \includegraphics[width=\linewidth]{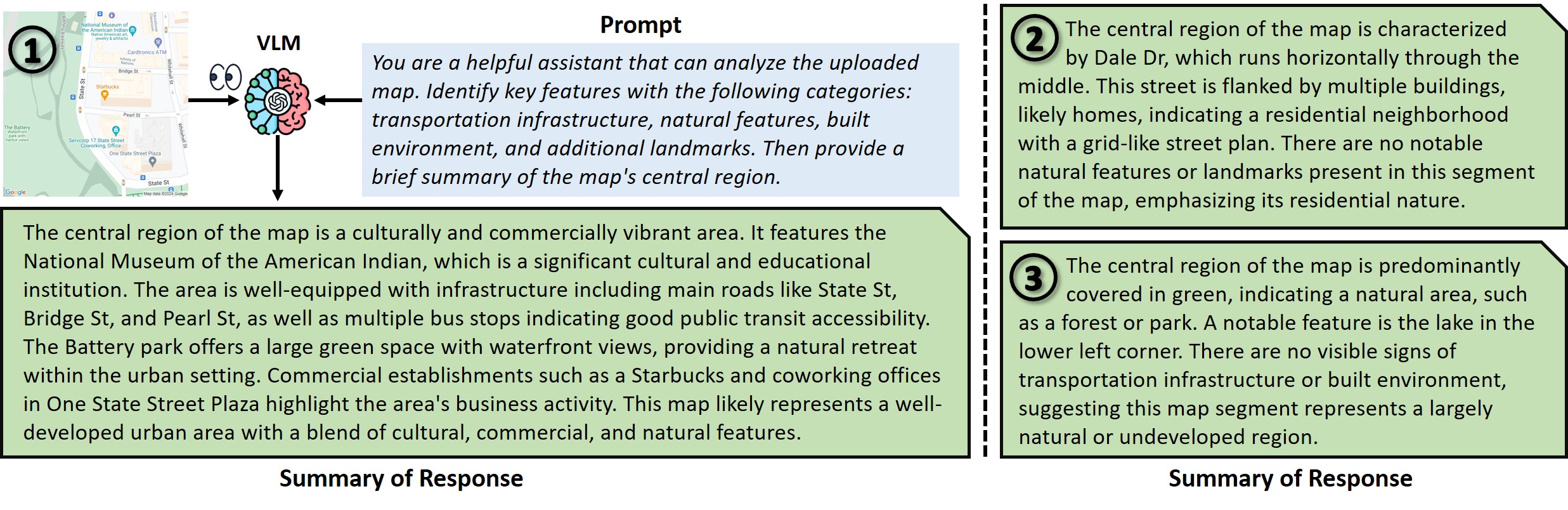}
  \caption{Examples of detecting location contexts by analyzing map images with VLM. The results are generated from GPT-4o \cite{OpenAI2024GPT4o} and input images are the maps in Figure \ref{fig:location:other}.}
  \label{fig:location:map}
\end{figure*}
Modern smartphones can easily access geographic locations, including latitude and longitude, through their positioning modules. However, GPS locations often do not provide sufficient information on their own. Our first idea is to exploit these locations with the existing Geographic Information Systems (GIS) like Google Maps \cite{GoogleMaps} or OpenStreetMap \cite{OpenStreetMap}, which offer comprehensive details about places worldwide and are widely used in daily life. However, identifying the location contexts from existing GIS is non-trivial. We first explore two available APIs of these GIS platforms:
\begin{itemize}
    \item \textbf{Reverse Geocoding API} \cite{GoogleMapsReverseGeocoding,nominatim}: This API converts geographic coordinates into addresses, providing a basic level of location context, such as `South Ferry, New York, NY 10004'.
    \item \textbf{Places API} \cite{GoogleMapsNearbySearch}: This API generates a list of nearby places within a specified radius around a geographic coordinate. It is important to note that there is a maximum limit on the number of place results, such as 20 for the Google Maps Places API \cite{GoogleMapsNearbySearch}.
\end{itemize}

We illustrate three example results of the above methods in Figure \ref{fig:location:other}, which cover different scenarios, including public, residential, and recreational areas. The addresses obtained from the Reverse Geocoding API do not convey informative location context. While the Places API can provide extensive landmarks information in urban areas like the `New York Stock Exchange' and `Charging Bull', it has limitations. In residential areas, as shown in the second example, the Places API tends to be biased toward public places, such as sports or educational centers, and may not accurately reflect the residential context. Furthermore, when the device is in a suburban area, both APIs may fail to return any relevant context. In summary, these two methods are not universally effective for location context detection across all scenarios.

To address this challenge, we observe that map segments on the other hand can provide more general and stable information than address or place texts. A map itself is an image where shapes, colors, and patterns all convey significant contextual information. For instance, the grey rectangles in the second case of Figure \ref{fig:location:other} likely represent houses, while the blue area in the third case indicates a body of water. More importantly, map segments are widely available and can be easily accessed through services like the Google Maps Static API \cite{GoogleMapsStaticAPI}. Therefore, we propose analyzing map images to derive more comprehensive location contexts.

However, interpreting maps is challenging, as it requires extensive knowledge to understand the shapes, colors, and texts presented in the images. Inspired by the rapid progress and success of recent vision language models (VLMs) \cite{driess2023palm,jiang2022vima,brohan2023rt,wang2023visionllm, OpenAI2024GPT4o}, we propose leveraging existing VLMs to analyze map images without any additional training. As shown in Figure \ref{fig:location:map}, we use GPT-4o \cite{OpenAI2024GPT4o} to detect location contexts from the three maps in Figure \ref{fig:location:other}. The results demonstrate GPT-4o's strong zero-shot ability to extract key features from the maps and generate accurate contexts for all three cases. Thus, modern VLMs offer a new and reliable approach to identifying location contexts from maps.

We use the Google Static Map API \cite{GoogleMapsStaticAPI} to retrieve map images, configuring three key parameters: the central location of the map (specified by geographic coordinates from the positioning module), the image size (500×500 pixels), and the zoom level (18), which ensures the map covers a sufficient area encompassing approximately 250×250 $m^2$ \cite{GoogleMapsStaticAPI}. To avoid redundant API calls for maps with close centers, we implement a grid system with a size of 100×100 $m^2$ and all coordinates in the same grid share the map image and location contexts. Additionally, since map information is generally stable, we maintain a key-value database to store the location contexts generated by VLMs. The key is a string representing the grid location, while the value is a string containing the location context. This approach allows us to reuse the inference results from VLMs, further reducing costs.



\subsubsection{Location Context from WiFi SSID}
\begin{figure}[t!]
  \centering
  \includegraphics[width=1.00\linewidth]{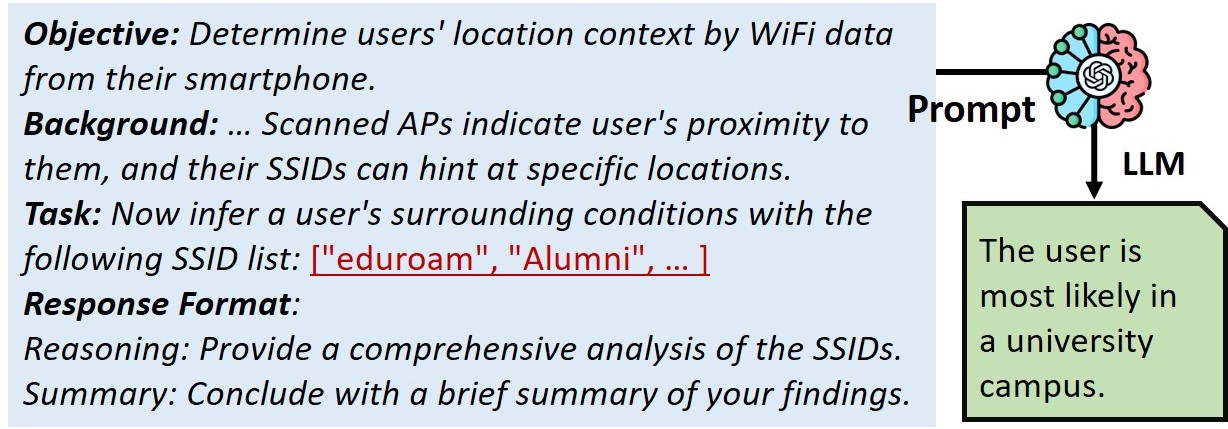}
  \caption{Location context detection with WiFi SSID. The red underlined texts in the prompt would be replaced by the scanned WiFi SSIDs.}
  \label{fig:location_wifi}
  \vspace{-0.5cm}
\end{figure}

In addition to GPS locations, WiFi Service Set Identifiers (SSIDs) can also provide valuable location context \cite{ni2022experience, xu2023penetrative}. For example, if a smartphone detects an SSID containing `Starbucks', it suggests that the user is near or inside a Starbucks. However, analyzing SSIDs requires a substantial amount of commonsense knowledge to interpret the names of various places, including restaurants, transportation hubs, landmarks, and more. To address this, we adapt the approach from \cite{xu2023penetrative} and utilize LLMs, such as ChatGPT \cite{openai2023gpt4}, to derive location contexts from WiFi SSIDs as shown in Figure \ref{fig:location_wifi}. We observe that many WiFi access points in public networks share identical SSIDs, such as `eduroam'. To optimize token usage, we preprocess the SSID list by removing duplicate SSIDs.

%


\subsection{Location Context Evaluation}
We conduct two experiments to evaluate the performance of existing commercial LLMs/VLMs in location context detection. The data collection process is detailed in Section 6. We find these tasks are special as analyzing maps or WiFi SSIDs requires a broad base of general knowledge, an area where existing LLMs may often outperform humans \cite{openai2023generalization}. To assess their performance, we evaluate the models by judging or rating their responses. We recruited 18 volunteers and collected a total of 330 and 360 scores for the two tasks, respectively.

In the first task of map interpretation using VLMs, we evaluate the performance of GPT-4o (\texttt{gpt-4o-2024-05-13}) \cite{OpenAI2024GPT4o}, Gemini Flash (\texttt{gemini-1.5-flash}) \cite{team2023gemini}, and Claude 3 Sonnet (\texttt{claude-3-5-sonnet-20240620}) \cite{anthropic2023claude}. We instruct the VLMs to generate descriptions for maps and designed a questionnaire to rate these descriptions. Each question included one map image, a description generated by an LLM, and four rating options ranging from 1 to 4, where `1' indicates "The description mismatches the map" and `4' represents "The description well matches the map". The questions were randomly sampled from 300 instances of map segments in Hong Kong, and the models were anonymized to the volunteers.  
\begin{figure}[t!]
  \centering
  \includegraphics[width=1.00\linewidth]{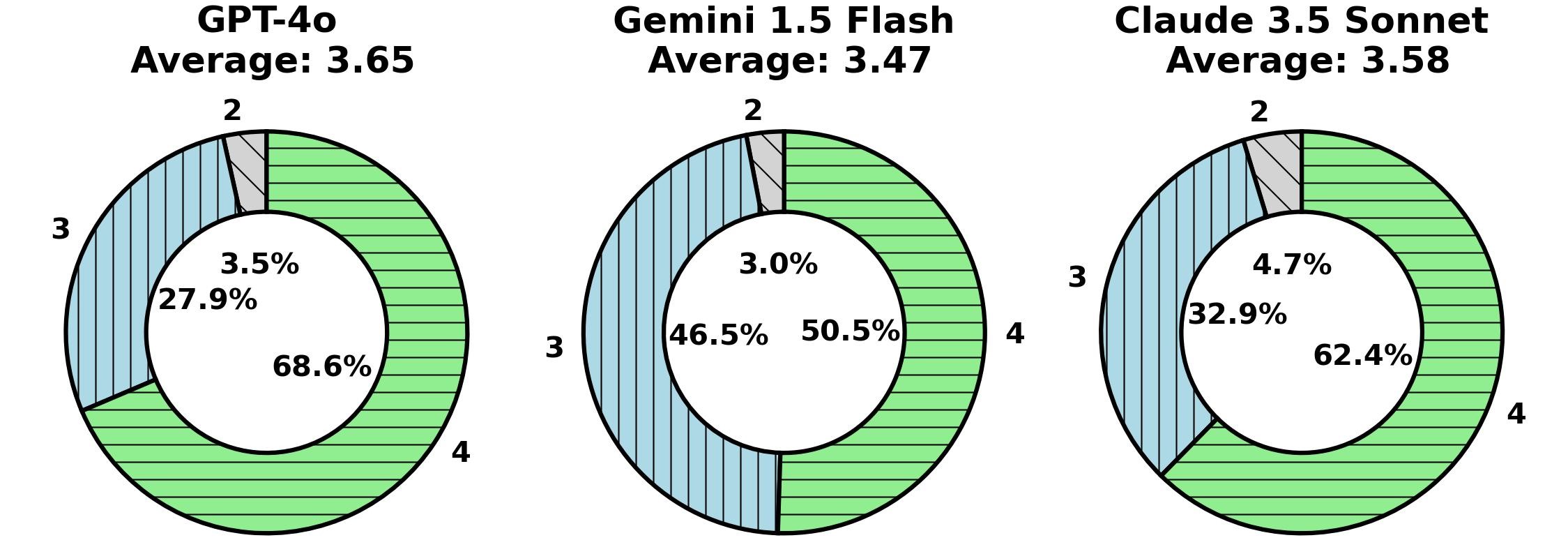}
  \caption{Performance of VLMs on location context detection with maps. The higher the scores, the better the performance.}
  \label{fig:loc:eva:map}
  \vspace{-0.5cm}
\end{figure}

Figure \ref{fig:loc:eva:map} presents the overall scores of the three VLMs that demonstrate impressive performance in this task, which requires interpreting shapes and texts (both in English and Chinese). The average scores were high, with GPT-4o, Gemini Flash, and Claude 3 Sonnet achieving 3.68, 3.47, and 3.58, respectively. Notably, none of the models hallucinates and receives a score of 1, underscoring the feasibility of using VLMs to interpret maps for location context detection.


\begin{table}[t!]
\centering
\caption{Performance of LLMs on location context detection with WiFi SSIDs. }
\scalebox{0.85}{
\begin{tabular}{lcccc}
\toprule
Metric &  \quad GPT-3.5 \quad & Gemini 1.5 Flash      & Claude 3 Sonnet       \\ \midrule 
Score \ ($\uparrow$)      & 3.51      & 3.43      &   3.25      \\ \midrule %
Win Rate\ ($\uparrow$)      & \textbf{42.2\%}      & 31.8\%      &   26.0\%      \\ \midrule %
Recall\ ($\uparrow$)      & 0.928         & 0.962       &   \textbf{0.997}   \\   
Specificity\ ($\uparrow$) & \textbf{0.895}         & 0.842     &  0.789     \\ \bottomrule 
\end{tabular}
}
\label{tab:loc:eva:wifi}
\end{table}

The second task, location context detection using WiFi SSIDs, is considerably more challenging for humans, as SSIDs often contain diverse and unfamiliar text, such as restaurant, company, or place names. We conducted 50 tests where volunteers rated the performance of LLMs on a scale from 1 to 4, with the assistance of ground-truth location context. For the remaining 310 tests, we had the LLMs compete against each other, asking volunteers to select the best response among. We also introduced two additional options: "SSIDs are not informative"—when SSIDs lack unique identifiers for detailed location contexts, and "Not sure"—when the models gives similar responses or when the SSIDs were particularly difficult to analyze.  Since this task involves only processing text inputs, we replaced GPT-4o (\texttt{gpt-4o-2024-05-13}) with lighter-weight GPT-3.5 (\texttt{gpt-3.5-turbo-0125}). 

\begin{figure*}[t!]
  \centering
  \includegraphics[width=0.8\linewidth]{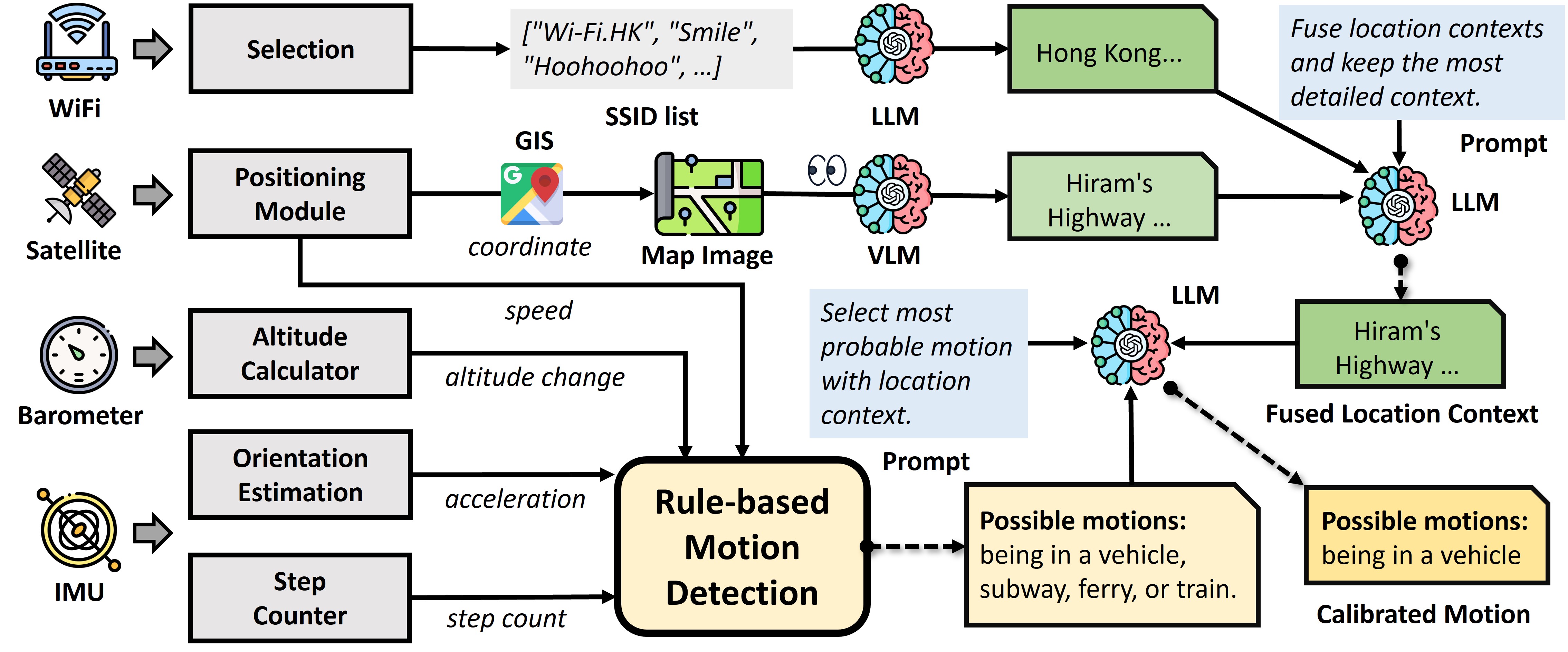}
  \caption{Location and motion context fusion in \sysname.}
  \label{fig:context:fusion}
  \vspace{-0.2cm}
\end{figure*}

Table \ref{tab:loc:eva:wifi} presents the performance of the three models across 360 tests. In this task, recall refers to the ratio of instances where the LLMs successfully generate valid context relative to the instances where volunteers consider SSIDs to be informative. Specificity represents the ratio of instances where LLMs generate valid context relative to the instances where volunteers believe SSIDs lack location indicators. Win rates indicate the number of cases in which each model beats the other two. Overall, all models achieve good performance, demonstrating that using them to analyze SSIDs for location context detection is effective.


\section{Context Fusion}\label{sec:con:fusion}
Now we have explored how to detect users' contexts with various sensors and this section will elaborate on how these contexts can be fused to enhance precision.

\subsection{Location Context Fusion}\label{sec:location:fusion}
Both map-based and SSID-based methods can provide valuable location contexts; however, we observe they have distinct features:
\begin{itemize}
    \item \textbf{Map-based location context} is effective in almost all situations but tends to provide only general descriptions, such as identifying an area as commercial or residential. Additionally, it struggles to offer detailed information in public areas with numerous points of interest (POIs); for example, it may not determine which specific store a user is in within a shopping mall.
    \item \textbf{SSID-based location context} can be fine-grained in some cases, such as identifying specific restaurants or campuses. However, it becomes less effective in suburban areas with few WiFi access points or when scanned SSIDs are not informative, such as `Redmi 9A' or `SjFaHJ6echEs,' which lack identifiers that can be used to derive meaningful location contexts.
\end{itemize}

Therefore, we propose fusing the two location contexts to obtain the most fine-grained context. Since both contexts are represented as text and the fusion task requires extensive commonsense knowledge, we believe LLMs are well-suited for this task. The upper part of Figure \ref{fig:context:fusion} illustrates the workflow for location context detection. The LLM is prompted to merge the location contexts and retain the most detailed and specific information—in the shown case, "Hiram's Highway", as nearby SSIDs are not highly informative. If the user is in an urban area, the SSID-derived context can provide valuable information, such as identifying a restaurant by an SSID like "McDonald's". This approach allows us to generate the most detailed and fine-grained location contexts based on multiple smartphone sensor signals.

\subsection{Motion Calibration}\label{sec:motion:location}
With the location context, actually we can further improve the accuracy of motion contexts, especially when our rule-based method provides multiple possible options. For instance, if a user is detected at a high GPS speed, determining the exact transportation mode can be challenging. But if we know the user is on a water surface, it's likely they are on a ferry. To achieve this, we propose calibrating the detected motion types using location context. 

This task also requires a significant amount of commonsense knowledge, making LLMs an effective solution. We represent both the location and motion contexts as text and use LLMs to calibrate the motions, as illustrated in Figure \ref{fig:context:fusion}. The LLM is prompted to "select the most probable motion given the location context". For example, if the primary location context is "Hiram's Highway", the transportation mode is likely to be "being in a vehicle". This approach allows us to further remove the ambiguity of motions and enhance the precision of motion contexts.

\section{Life Journaling}
The previous section details how to obtain accurate contexts from sensor data, though this process is limited to short time windows, e.g., 15 seconds. But generating a life journal requires processing sensor data over much longer durations like hours. This section explains how to aggregate contexts from extended time windows and generate life journals. 

\subsection{Context Refinement}\label{sec:con:ref}
To get long-term context information, we should aggregate context logs over time. However, simply combining these contexts as texts can result in overly lengthy and less accurate data. To address this, we apply several optimizations to the context fusion process.

First, we observe that location contexts from neighboring time windows may vary in quality or detail. For example, one context might describe "a restaurant", while the context from the neighboring window can specify "a McDonald's restaurant", with the latter providing more information. Therefore, we also need to fuse location contexts over time. Additionally, as shown in Figure \ref{fig:location:map}, the location contexts generated by LLMs are often lengthy. For instance, the token size of the location context in Figure \ref{fig:location:map} case 1 is 131 tokens for ChatGPT. Directly aggregating this length of context over an hour would result in a text with a token size of 7,860 if we detect map-based location context every minute. To reduce the text length, we introduce a simple yet effective instruction in the prompt when fusing location contexts derived from maps and SSIDs, i.e., "present concise location logs" or "present concise motion logs".
\begin{figure}[t!]
  \centering
  \includegraphics[width=0.90\linewidth]{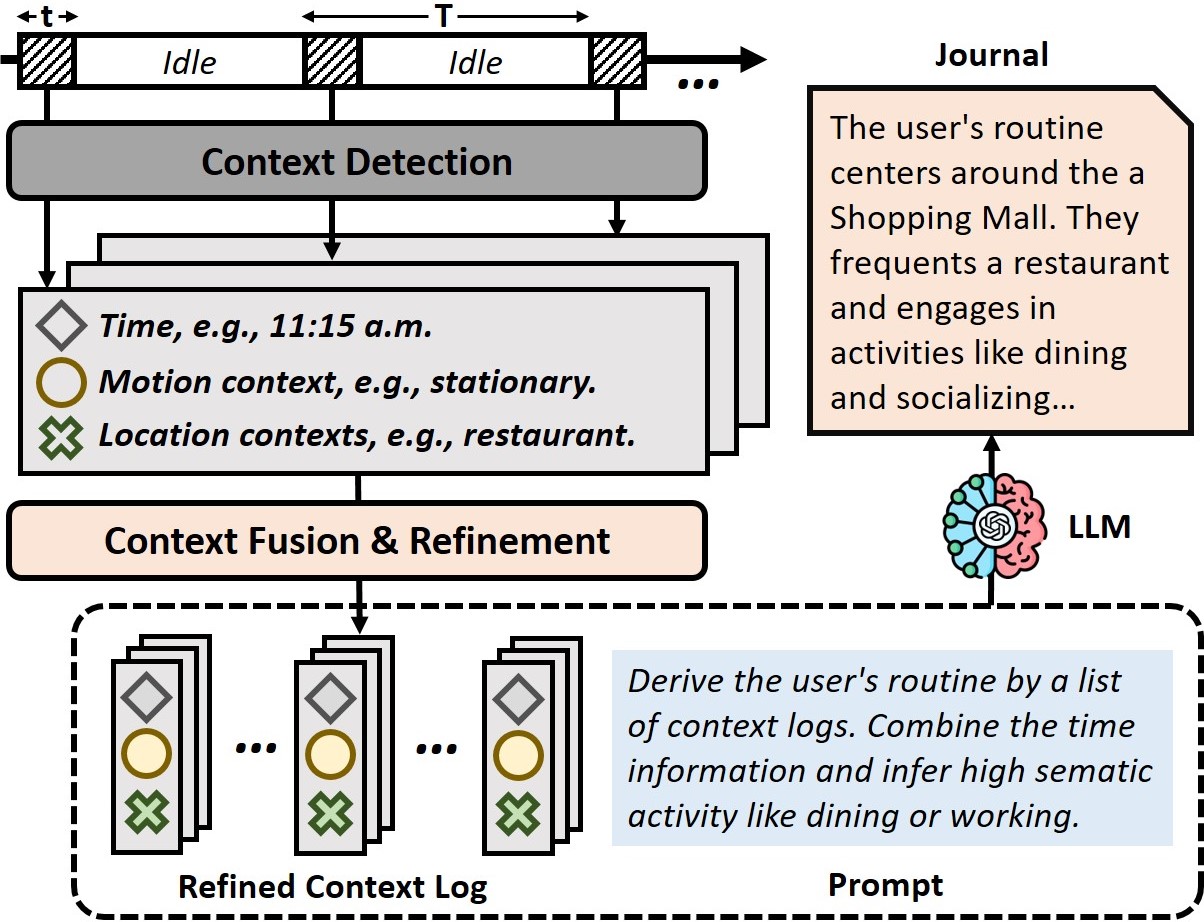}
  \caption{Journal generation in \sysname.}
  \label{fig:journal}
  \vspace{-0.55cm}
\end{figure}

Combining all the designs, we organize multiple location contexts from neighboring time windows in the format of "\textit{[time-1](map location context, WiFi location context), ..., [time-n](map location context, WiFi location context)}" to further incorporate time context, where $n$ is set to 15 in \sysname. The LLM is instructed to perform three steps: 1) "select the most detailed location context from two contexts at the same time", 2) "select the most specific and detailed location context across time", and 3) "present the enhanced and concise location logs as \textit{[time-1](fused location context), ..., [time-n](fused location context)}". The refined location contexts are extracted and then used to calibrate the motion contexts as outlined in Figure \ref{fig:context:fusion}.

\subsection{Journal Generation}\label{sec:jou:gen}
Now we can combine these refined contexts to cover longer durations like hours. We organize the three contexts over time as "\textit{[time-1](calibrated motion context, fused location context), ..., [time-n](calibrated motion context, fused location context)}". Similarly, we believe that the task of deriving a journal from a list of contexts is well-suited for LLMs, as it requires a substantial amount of common sense knowledge. As shown in Figure \ref{fig:journal}, we provide the LLMs with a prompt that instructs them to analyze the context logs and infer high-level semantic activities like dining. To improve the journal quality and control the format of the generated journals, we also include several example journal entries in the prompt, such as, "In the morning, the user spends time at a local library, likely reading and researching".

We also observed that many LLMs, like ChatGPT, tend to include "subjective comments" on the response, such as, "The routine consists of a blend of work and leisure". To address this, we use another LLM session with the prompt -"remove any subjective comments if they exist" to further polish the journal. This process yields the final journal for the user, summarizing their behaviors over a long duration.

\subsection{Data Collection Duty Cycle}\label{sec:journal:duty}
Although life journaling requires long durations of sensor data, it is unnecessary for our system to continuously and consistently collect data from smartphones, such as scanning WiFi signals for hours, as this would consume excessive energy \cite{androidWiFiScan}. Therefore, we design a duty cycle for the data collection, as shown at the top of Figure \ref{fig:journal}. The system periodically activates the collection process and then enters an idle state for a while. The context detection module then processes the collected sensor data to generate contexts. The parameters $t$ and $T$ represent the collection duration and period, respectively. To allow sufficient time for the smartphone to scan WiFi and compute a more accurate step count, we set $t$ to 15 seconds. The collection period $T$ is set to 60 seconds and its impact will be evaluated in Section \ref{sec:eva:sam}.
\section{Evaluation}
\begin{figure*}[t!]
  \centering
  \includegraphics[width=0.95\linewidth]{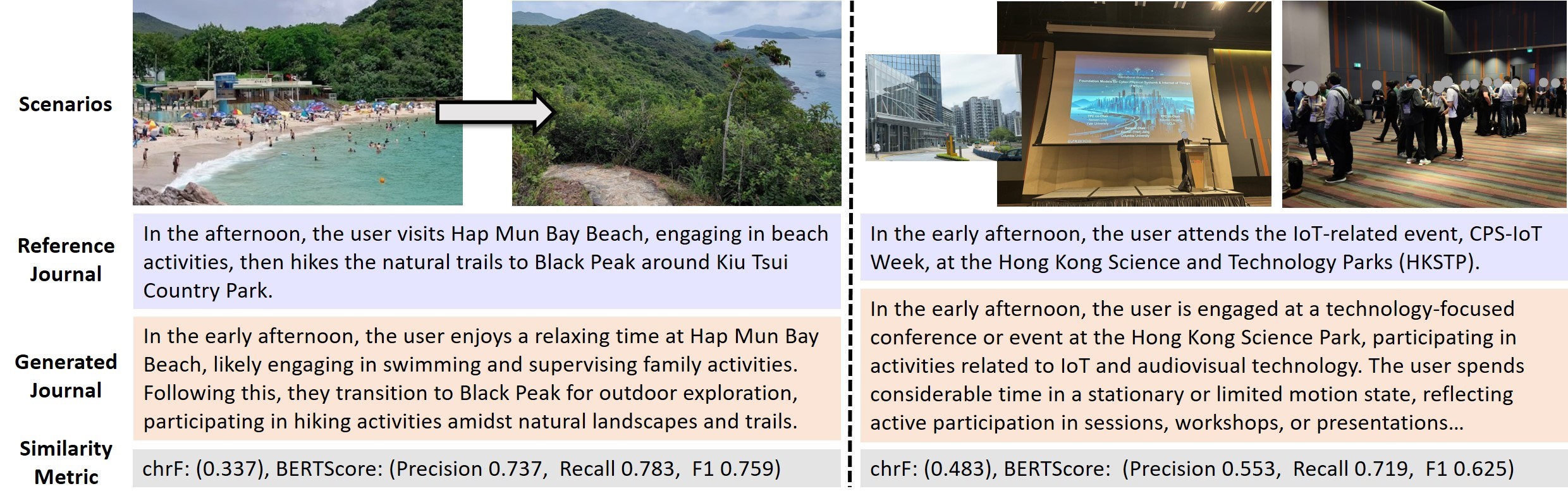}
  \caption{Life journal examples generated by \sysname with GPT-4o mini.}
  \label{fig:eva:showcase}
\end{figure*}

\subsection{Implementation}\label{sec:imp}

\textbf{APP design}. Since life journaling is a novel application, to the best of our knowledge, there is no existing dataset available for it. Therefore, we develop an Android application that runs a foreground service to regularly access sensor data, such as satellite and WiFi signals, from the system APIs. The data collection process follows the duty cycle described in Section \ref{sec:journal:duty}, with all sensor data being implicitly saved in files for offline propcessing. 


\textbf{Dataset}. We recruit 4 volunteers from Hong Kong to collect an extensive dataset in various scenarios with three smartphones including Samsung Galaxy S8, Samsung Galaxy S22, and Google Pixel 7. During the data collection process, each volunteer carries the experimental smartphone and goes about their daily activities as usual, activating the data collection in the application. The smartphone was not required to be tightly attached to the volunteers; for example, they were free to place the phone on a table while having a meal. We collect data from 58 experiments, totaling 4,417 minutes, with an average experiment duration of 76.2 minutes — significantly longer than the sensing durations, e.g., typically a few seconds, used in HAR studies. For each experiment, the corresponding volunteer provides two similar and concise text descriptions of their behaviors, referred to as the \textit{reference journals}, for evaluation purposes.


\textbf{Models}. \sysname has many LLM-based modules, and there are numerous potential combinations of available models. We establish a default configuration with several representative LLMs. In the location context module, we select GPT-4o (\texttt{gpt-4o-2024-05-13}) \cite{OpenAI2024GPT4o} for map interpretation and GPT-3.5 (\texttt{gpt-3.5-turbo-0125}) \cite{openai2023gpt4} for SSID interpretation. In the context fusion and journal generation modules, we adopt GPT-4o mini (\texttt{gpt-4o-mini-2024-07-18}).

\textbf{Prompts}. All prompts include a specified response format ("reasoning" and "summary" \cite{wei2022chain}) to constrain LLMs, and every response undergoes a keyword detection process to extract the key "summary" content, such as location context or journal entries. Due to space constraints, we illustrate key parts of the prompts in Figure \ref{fig:location:map}, \ref{fig:context:fusion}, and \ref{fig:journal}.

\textbf{Metrics}. To evaluate the quality of journals generated by \sysname, we measure the similarities between them and the reference journals using chrF \cite{popovic2015chrf} and BERTScore \citep{zhang2019bertscore}, both of which are widely adopted metrics in the natural language processing domain. We also define LLMs as hallucinating if they do not follow the specified response format and the target context cannot be extracted from their responses.

\textbf{Baseline}. To comprehensively evaluate \sysname, we also establish a baseline solution, referred to as \textit{SenLLM} in this paper, which simply aggregates raw sensor across time and inputs them into LLMs for journal generation.

\begin{table}[t!]
    \centering
    \caption{Performance of \sysname and baseline methods. `Hall.' denotes hallucination while `P', `R', and `F1' represents precision, recall, and F1 score, respectively.}
    \scalebox{0.88}{
    \renewcommand{\arraystretch}{1.1}
    \begin{tabular}{l|c|ccccc}
        \toprule
        \multirow{2}{*}{LLM}& \multirow{2}{*}{Method} & \multirow{2}{*}{\begin{tabular}[c]{@{}c@{}}Hall.\\ rate ($\downarrow$) \end{tabular}} & \multirow{2}{*}{chrF ($\uparrow$)}  & \multicolumn{3}{c}{BERTScore ($\uparrow$)} \\ \cline{5-7} 
        &  & & & P  & R & F1 \\ \midrule
        \multirow{2}{*}{GPT-4o} 
        & SenLLM & 0.000 & 0.451 & 0.592 & 0.680 & 0.630\\
        & \sysname & 0.000 & 0.509 & 0.613 & 0.772 & 0.681\\ \midrule
        \multirow{2}{*}{\begin{tabular}[l]{@{}l@{}}GPT-4o\\ mini\end{tabular}} 
        & SenLLM & 0.000 & 0.394 & 0.563 & 0.621 & 0.588\\
        & \sysname & 0.000 & \textbf{0.553} & 0.641 & 0.776 & 0.699\\ \midrule
        \multirow{2}{*}{\begin{tabular}[l]{@{}l@{}}Claude\\ 3 Opus\end{tabular}} 
        & SenLLM & 0.000 & 0.437 & 0.622 & 0.692 & 0.652\\
        & \sysname & 0.000 & 0.536 & 0.646 & \textbf{0.782} & \textbf{0.704}\\ \midrule
        \multirow{2}{*}{\begin{tabular}[l]{@{}l@{}}Gemini \\ 1.5 Pro\end{tabular}} 
        & SenLLM & 0.000 & 0.400 & 0.587 & 0.645 & 0.611\\
        & \sysname & 0.000 & 0.483 & 0.637 & 0.737 & 0.680\\ \midrule
        \multirow{2}{*}{\begin{tabular}[l]{@{}l@{}}Llama3 \\ 70B*\end{tabular} } 
        & SenLLM & 0.052 & 0.411 & 0.594 & 0.630 & 0.608\\
        & \sysname & 0.000 & 0.505 & \textbf{0.650} & 0.738 & 0.688\\
        \bottomrule
    \end{tabular}
    }
    \label{tab:eva:main}
    \vspace{-0.5cm}
\end{table}
\subsection{Main Results}
Figure \ref{fig:eva:showcase} shows two example journals generated by \sysname together with ground-truth scenario photos and reference journals. In the first case, the user visits a beach and then goes hiking. The \sysname successfully captures the key activity like `hiking' and location context like the name of the beach. Similarly, the generated journal also demonstrates high quality in the second case and \sysname derives the user attending a conference or event. Interestingly, it derives that event is IoT-related from a scanned SSID "CPS-IoT WEEK 2024". Overall, the generated journal aligns well with the reference journal and their similarities achieve high BERTScore.

Interestingly, we observe that LLMs sometimes give some complementary descriptions like "participation in sessions", which are valid but do not appear in the reference journals. Additionally, LLM can make some reasonable speculations based on the motion and location contexts, e.g., `swimming'. Both factors typically result in the generated journal being longer than the reference journal, which causes the recall to be higher than the precision. 

To provide a comprehensive quantitative evaluation, we test \sysname and the baseline solution SensorLLM using different LLMs for journal generation, including GPT-4o \cite{OpenAI2024GPT4o}, Claude 3 \cite{anthropic2024claude3}, Gemini 1.5 \cite{reid2024gemini}, and Llama3 \cite{meta2024llama3}. Table \ref{tab:eva:main} presents their overall performance across various metrics. Interestingly, we find that many LLMs within the SensorLLM can capture some insights about users' behaviors solely from WiFi SSIDs. However, Llama3 70B in SensorLLM shows a hallucination rate of 5.2\%, and most LLMs achieve lower scores. In contrast, LLMs integrated with \sysname achieve significantly better results; for example, Claude 3 Opus achieves high BERTScore precision and recall of 0.646 and 0.782, respectively. Notably, none of the models hallucinate during the challenging task of life journaling, and the much lighter-weight and open-source model like Llama3 70B also performs well. Overall, the results clearly demonstrate the superior effectiveness of \sysname over the baseline solution.

\subsection{Impact of Time Period}
\begin{figure}[t!]
\begin{subfigure}{.48\linewidth}
  \centering
  \includegraphics[width=\linewidth]{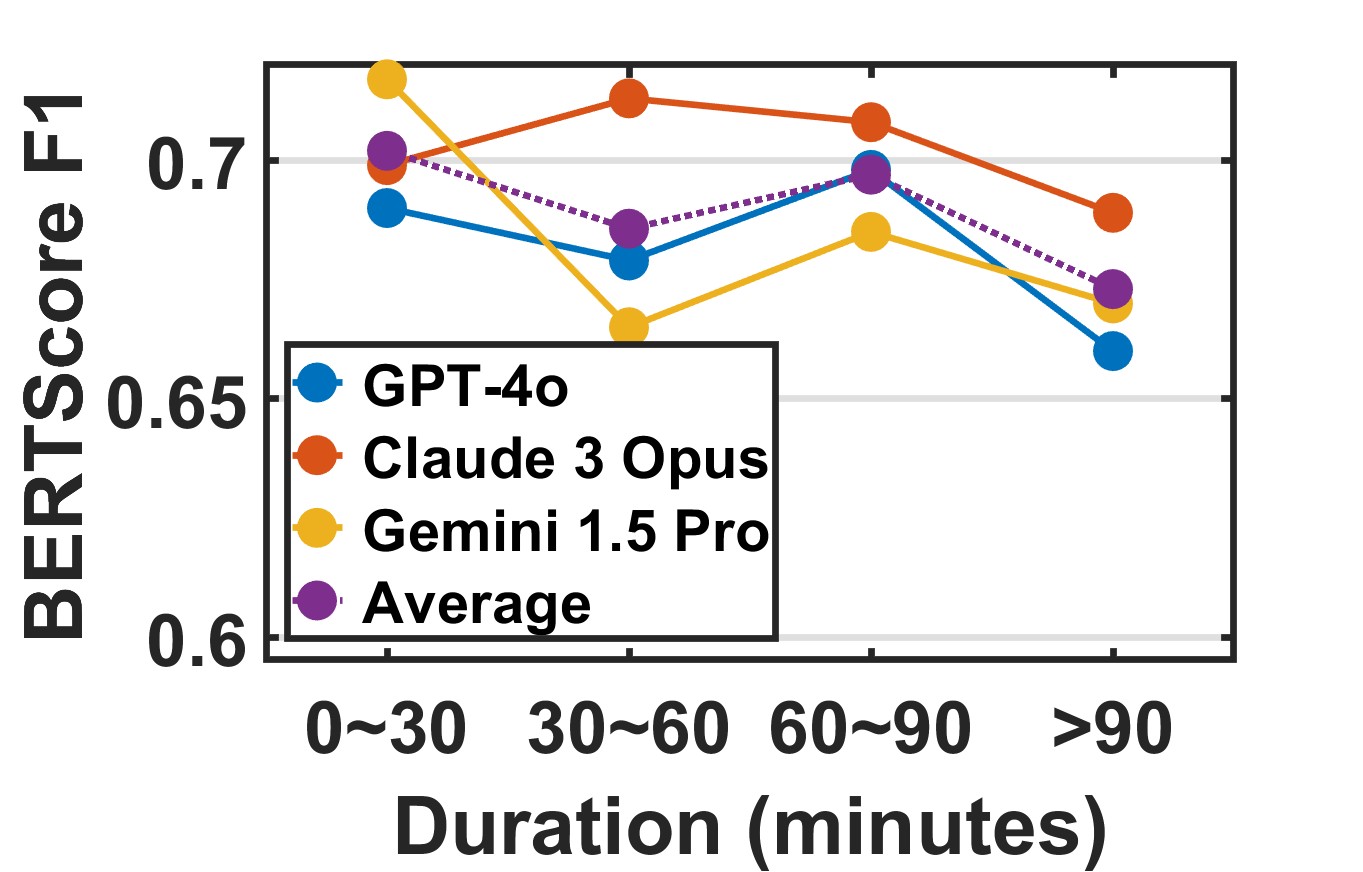}
  \caption{Journal duration.}
  \label{fig:eva:dur}
\end{subfigure}
\begin{subfigure}{.48\linewidth}
  \centering
  \includegraphics[width=\linewidth]{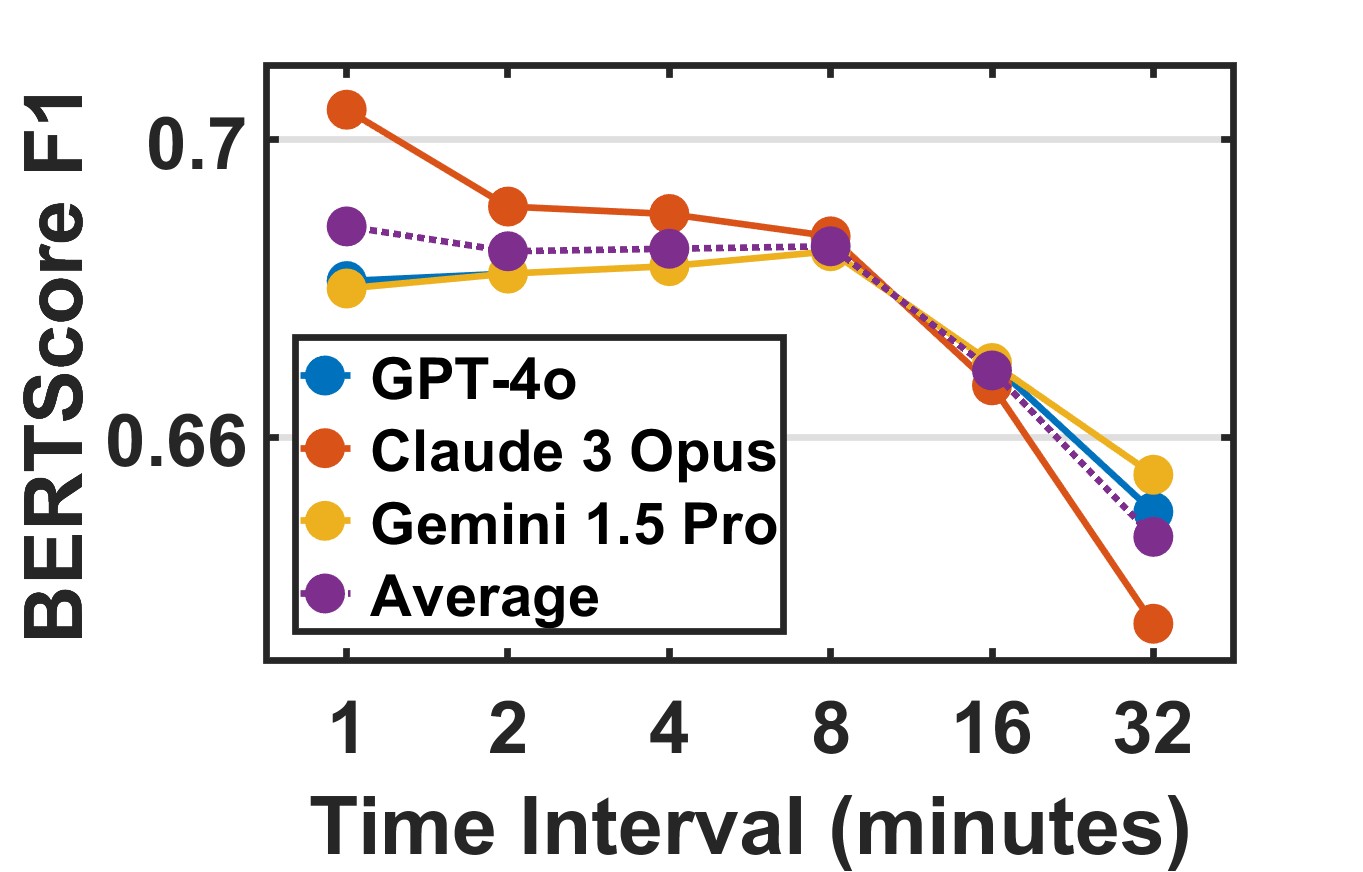}
  \caption{Sampling interval.}
  \label{fig:eva:sam}
\end{subfigure}
\caption{Performance of \sysname under different settings.}
\vspace{-0.5cm}
\end{figure}
We then examine the impact of experiment duration on \sysname, and Figure \ref{fig:eva:dur} presents the results for three representative LLMs across different durations, ranging from 0–30 minutes to over 90 minutes. As the duration increases, the LLMs maintain good performance, with only a slight overall decrease in BERTScore F1. Notably, all models achieved scores higher than 0.66, even for durations exceeding 90 minutes. These results indicate that \sysname is not highly sensitive to duration and can effectively generate journals for long-term sensor data, such as 90-minute windows.

\subsection{Impact of Sampling Interval}\label{sec:eva:sam}
As detailed in Section \ref{sec:journal:duty}, we designed a data collection duty cycle where the application periodically collects data from the smartphone. This experiment evaluates the impact of the sampling interval on the quality of generated journals. As shown in Figure \ref{fig:eva:sam}, the results show that all three models achieve stable and high BERTScore F1 when the intervals range from 1 to 8 minutes. However, when the interval increases over 16 minutes, the overall performance degrades significantly across all models. While a higher sampling interval reduces system overhead, such as power consumption and token usage, it also leads to information loss and lower-quality journals, making it a trade-off parameter.

\subsection{Ablation Study}
\begin{figure}[t!]
\begin{subfigure}{.48\linewidth}
  \centering
  \includegraphics[width=\linewidth]{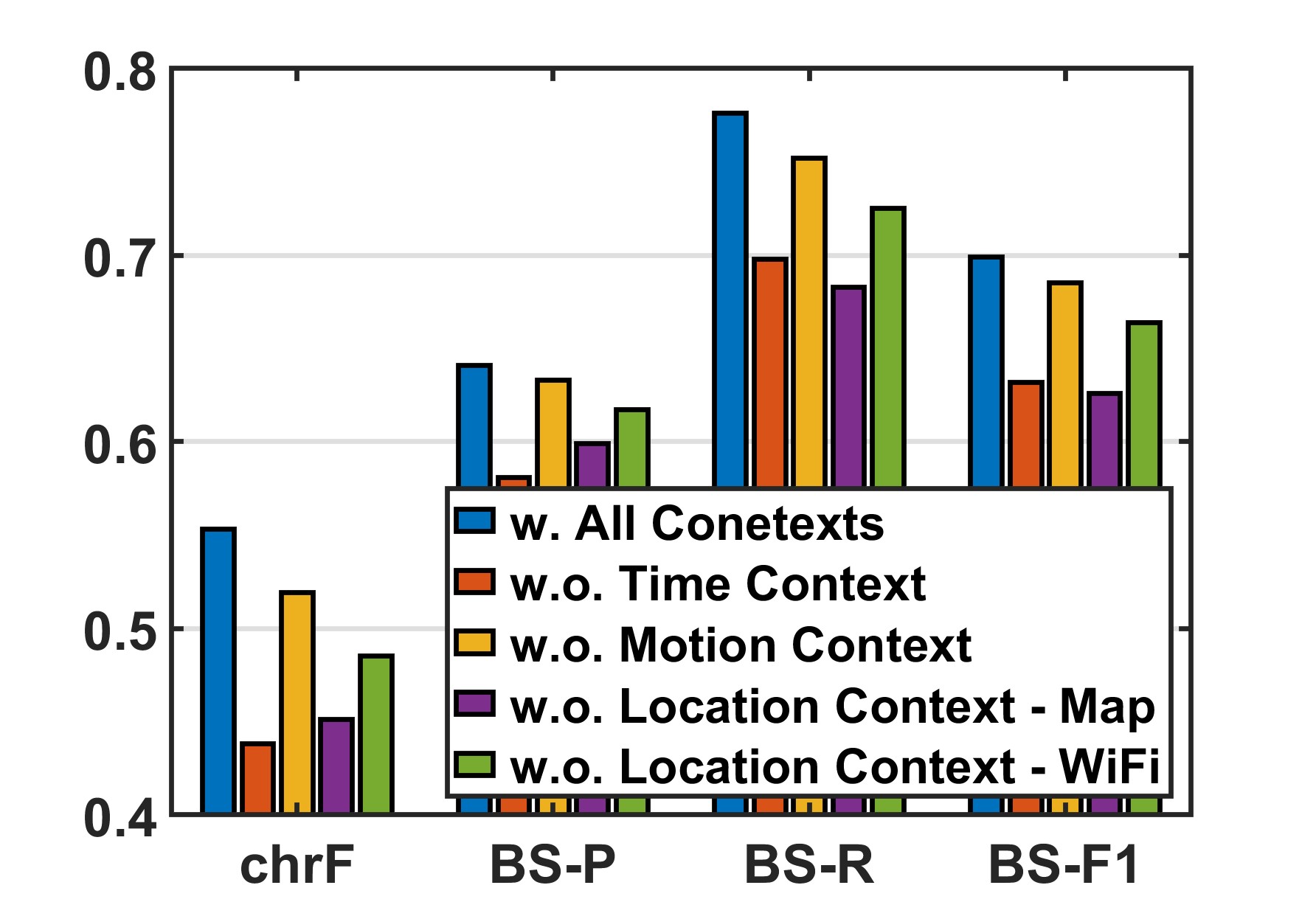}
  \caption{Varying data resources.}
  \label{fig:eva:res}
\end{subfigure}
\begin{subfigure}{.48\linewidth}
  \centering
  \includegraphics[width=\linewidth]{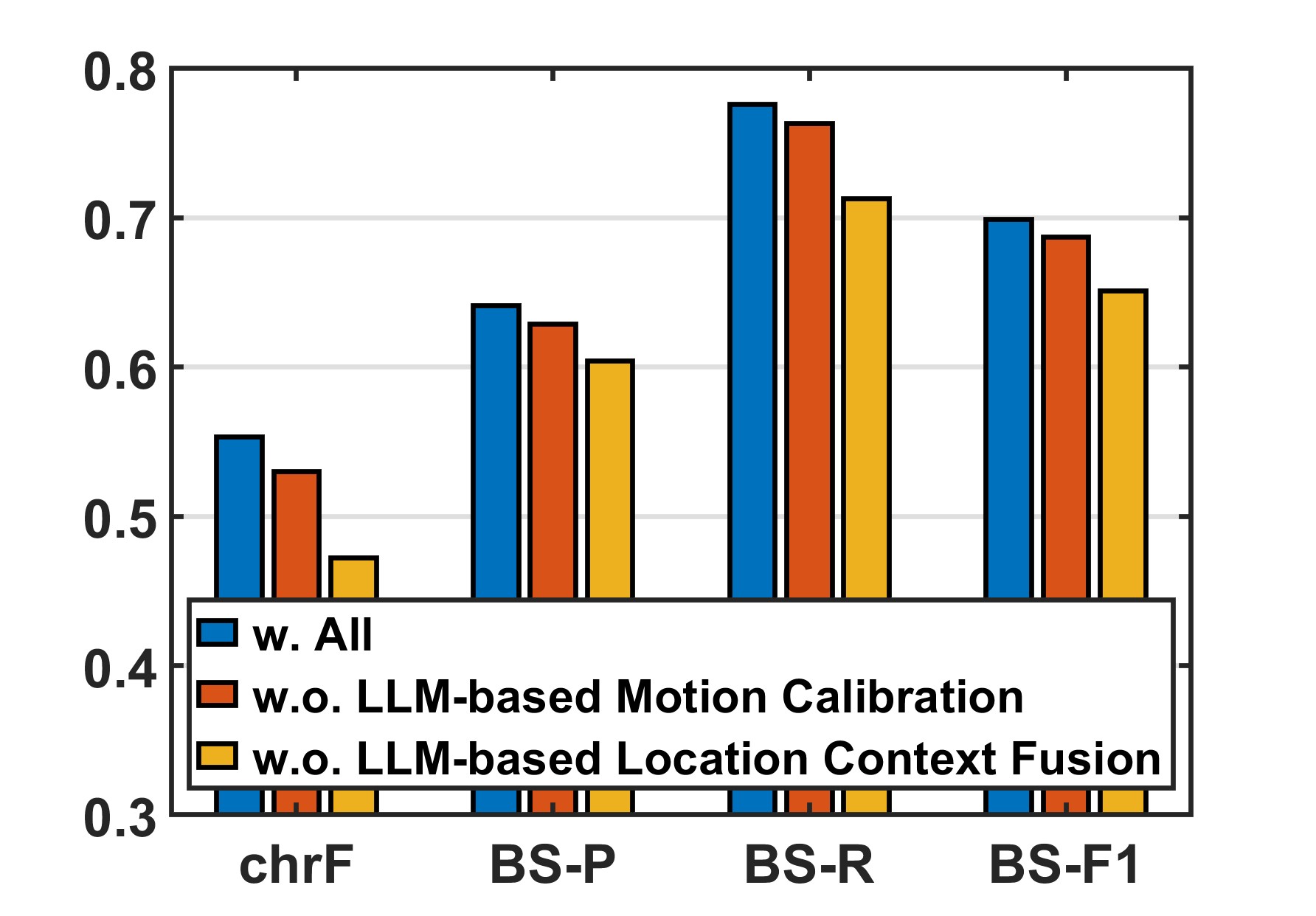}
  \caption{Varying context fusion.}
  \label{fig:eva:con}
\end{subfigure}\\
\begin{subfigure}{.48\linewidth}
  \centering
  \includegraphics[width=\linewidth]{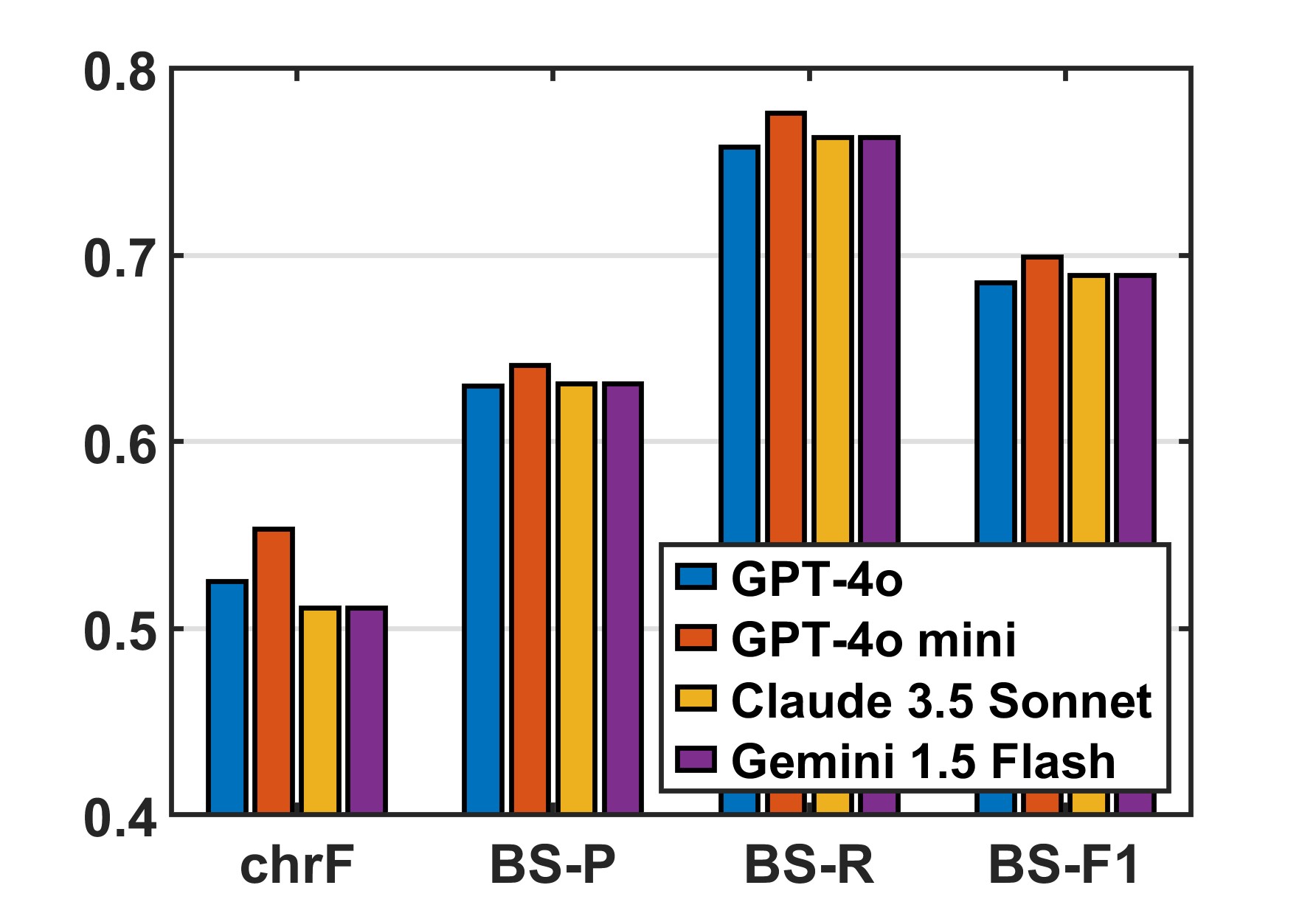}
  \caption{Varying fusion model.}
  \label{fig:eva:fmodel}
\end{subfigure}
\begin{subfigure}{.48\linewidth}
  \centering
  \includegraphics[width=\linewidth]{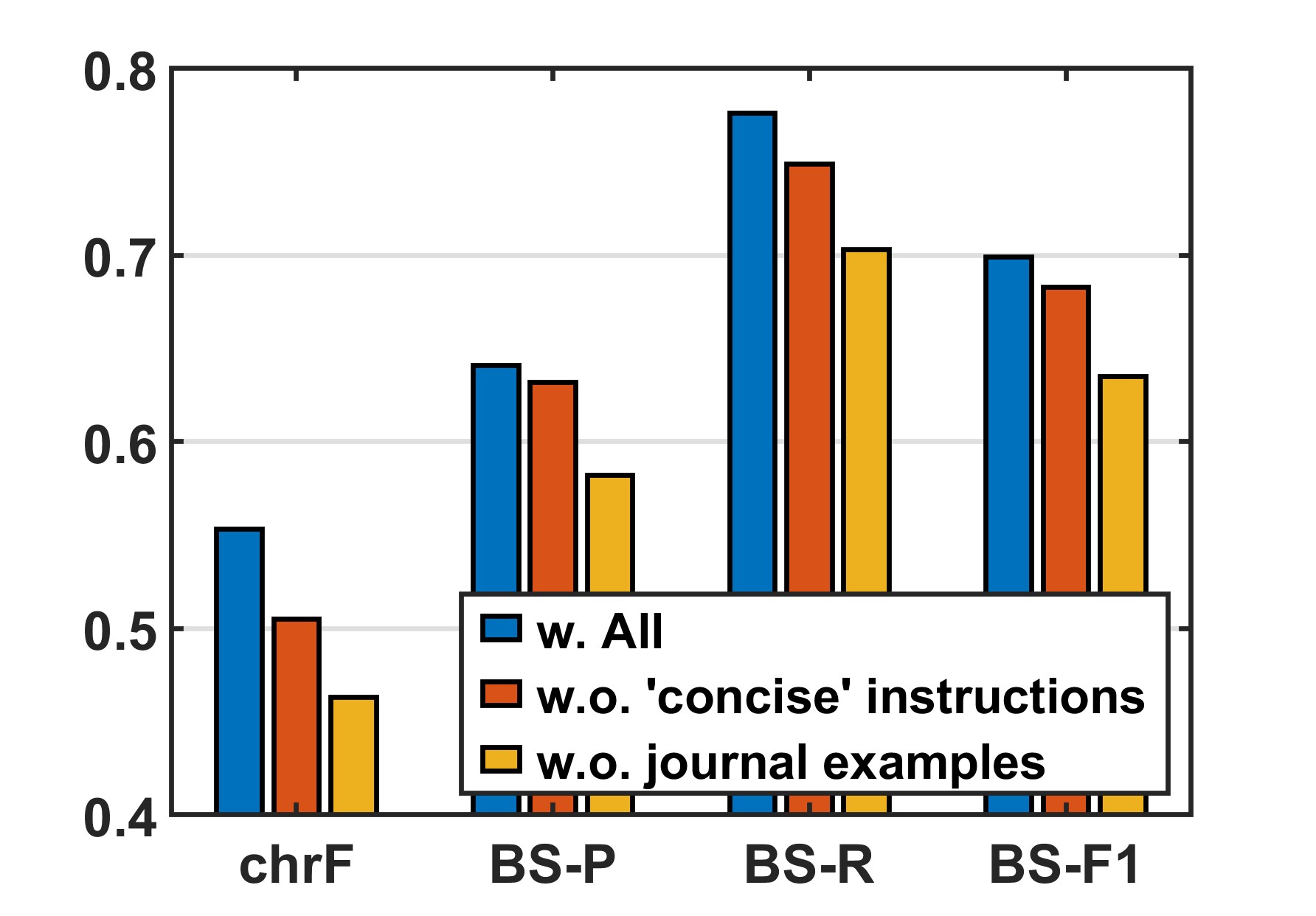}
  \caption{Varying prompt.}
  \label{fig:eva:prompt}
\end{subfigure}
\caption{Impact of different design components. The `BS-P', `BS-R', and `BS-F1' indicate the BERTScore precision, recall, and F1, respectively.}
\vspace{-0.5cm}
\end{figure}



\textbf{Impact of resources}. We first investigate the impact of different data sources on journal generation. Figure \ref{fig:eva:res} shows the quality of journals generated using various combinations of resources. For example, `w.o. motion' indicates that only the location context was used for journal generation. The results demonstrate that combining all available resources—including both motion and location contexts (map-based and WiFi-based)—yields the best performance for \sysname. Notably, the map location context plays a crucial role in journal quality. Removing it resulted in a decrease of 0.042, 0.093, and 0.073 for BERTScore precision, recall, and F1, respectively.

\textbf{Impact of context fusion}. We also evaluate the performance of \sysname without LLM-based location context fusion (Section \ref{sec:location:fusion}) or LLM-based motion calibration (Section \ref{sec:motion:location}). As shown in Figure \ref{fig:eva:con}, comparing \sysname with these two alternatives reveals that both contribute to improvements across the four metrics. For instance, omitting LLM-based location context fusion leads to a 0.081 decrease in chrF. These results show the effectiveness of LLM-based context fusion and enhanced contexts can benefit the downstream journal generation task.

\textbf{Impact of context fusion models}. Different from Table \ref{tab:eva:main}, this experiment focuses on the impact of LLMs on context fusion (Section \ref{sec:con:fusion}). As shown in Figure \ref{fig:eva:fmodel}, we test three representative LLMs for both fusing map- and WiFi-based location contexts and calibrating motions using the fused location context. Although these models are not high-end LLMs, they still achieve fair performance, demonstrating the effectiveness of \sysname's task decomposition, which allows LLMs to handle each subtask effectively. These results, along with those in Table \ref{tab:eva:main} confirm the generalizability of \sysname across different LLMs.

\textbf{Impact of prompt}. In Figure \ref{fig:eva:prompt}, we examine how two key designs in prompts impact journal qualities by removing the "present concise description" instruction for context refinement (Section \ref{sec:con:ref}) or the journal examples for the journal generation (Section \ref{sec:jou:gen}). The `concise' instruction provides a slight overall performance improvement while significantly reducing token usage, which will be discussed in the next subsection. Including journal examples, however, contributes to a more substantial improvement in overall performance.

\subsection{System Cost}
\begin{table}[t!]
\centering
\caption{Token usage summary of \sysname. The token usage includes both input and output token numbers.}
\scalebox{0.92}{
\begin{tabular}{lccc}
\toprule
Module                & Token Usage           & Freq.             & Price (dollar)                              \\ \midrule
Map Context           & 437, 316               & 1/min                 & $1.5\times10^{-2}$/hr                      \\
WiFi Context          & 309, 335               & 1/min                 & $1.5\times10^{-2}$/hr                      \\
Location Fusion       & 1236, 611              & 4/hr              & $5.9 \times 10^{-4}$/hr                      \\
Motion Calibration    & 602, 395               & 4/hr              & $3.5 \times 10^{-4}$/hr                      \\
Journal Generation    & 2015, 394              & 1/hr                & $5.4 \times 10^{-4}$/hr                      \\
Journal Clearning     & 98, 60                 & 1/hr                & $5.1 \times 10^{-5}$/hr                     \\ \midrule
Total & - & - & $ 3.2 \times 10^{-2}$/hr \\ 
\bottomrule
\end{tabular}}
\label{tab:eva:cost}
\vspace{-0.3cm}
\end{table}
We evaluate the overall system cost of \sysname using GPT-4o mini and pricing as of August 2024. The frequency of journal generation and cleaning (removing "subject" comments introduced in Section \ref{sec:jou:gen}) is set to once per hour and all token usages are the averages across all experiments. As shown in Table \ref{tab:eva:cost}, the total cost is $\$3.2 \times 10^{-2}$ per hour. Additionally, by adopting a map-based location context database, map contexts can be reused and the token usage can be reduced by 82\%, lowering the total cost to $\$2.2 \times 10^{-2}$ per hour. The token usages with the `concise' instruction are reduced by 5.1\%, 7.8\%, and 9.0\% for the outputs of location fusion, motion calibration, and the input for journal generation, respectively. Overall, the system cost is affordable using commercial LLMs, which can be further reduced by leveraging open-source models like Llama 3.

\subsection{User Study}
We also conducted a user study experiment to evaluate how the generated journals met the quality standards expected by users using five key metrics as follows: (1) \textbf{Clarity} is assessed by examining how easy the journal is to understand and whether the information is presented logically and coherently. (2) \textbf{Conciseness} evaluates whether the journal conveys its message efficiently, avoiding redundant information. (3) \textbf{Correctness} focuses on the accuracy of the content, measuring that the information presented is factual and error-free. (4) \textbf{Completeness} ensures that the journal thoroughly covers all relevant aspects, providing the necessary detail without omitting relevant information. (5) \textbf{Relevance} assesses the degree to which the content is focused on important and meaningful aspects.

\begin{figure}[t!]
  \centering
  \includegraphics[width=1.00\linewidth]{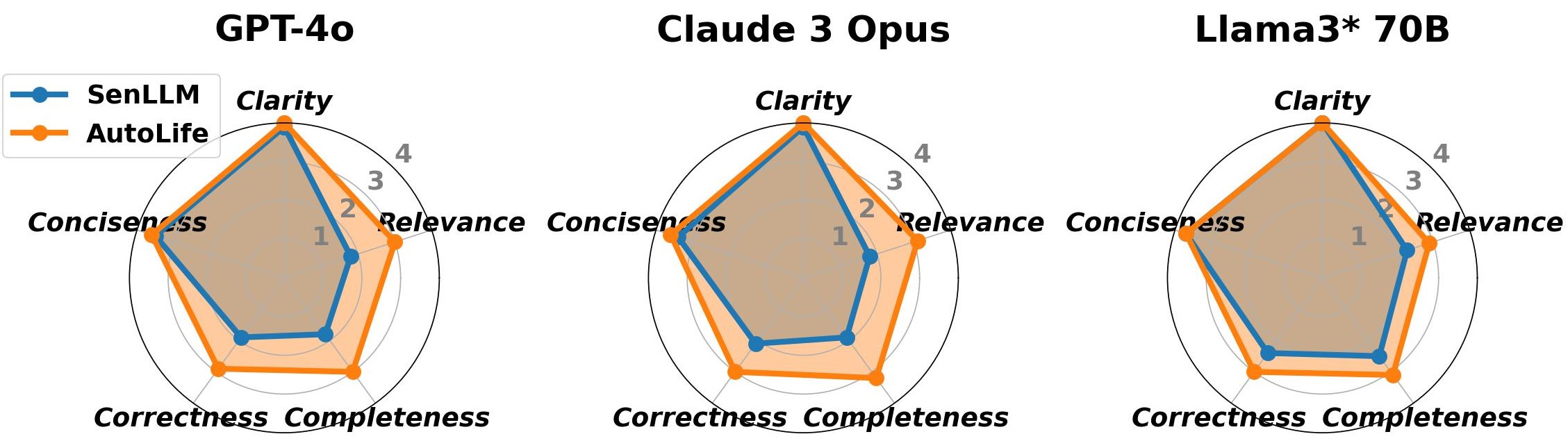}
  \caption{User study results for \sysname.}
  \label{fig:user:study}
  \vspace{-0.3cm}
\end{figure}

Eight volunteers rated each metric for randomly sampled 20 experiments on a four-point scale, where 1 indicates "completely does not meet the criteria", and 4 indicates "completely meets the criteria". Figure \ref{fig:user:study} shows the average scores of three models rated by volunteers, with \sysname achieving scores higher than 3.0 among most metrics, significantly outperforming SensorLLM in terms of correctness, completeness, and relevance. These results further validate the effectiveness and usability of \sysname.

\section{Discussion}
\textbf{Use cases}: Life journaling has the potential to enable a wide range of valuable downstream applications. For example, the system can create comprehensive, long-term memos that users can easily retrieve for reflection or reference. It can also automatically generate detailed travel logs for personal use or sharing on social media. Additionally, the system can produce time-use reports, such as "You spent 3 hours commuting and 5 hours in meetings today", helping users gain insights into their daily routines. By analyzing users' daily routines, we can develop more comprehensive user profiles to accurately recommend activities, products, or services to their preferences.

\textbf{Privacy Concerns}: Life journaling inherently involves handling sensitive data, raising significant privacy considerations. Our app implements the data collection module in a foreground service \cite{android_foreground_service}, which is visibly displayed as a notification bar and ensures that users are fully aware of and can monitor the data collection process. Future work like processing data locally on devices instead of cloud-based models effectively safeguards user privacy. Additionally, giving users full control over the collected sensor data and all generated outputs, including contexts and journals, can greatly enhance their sense of safety and trust in the system.



\section{Conclusion}
In this paper, we propose a novel mobile sensing application called \textit{life journaling} and design an automatic life journaling system \sysname utilizing ubiquitous smartphones. To accurately derive a user's journal, \sysname exploits multiple contexts and extensive common knowledge within LLMs. We collect a dataset and establish a benchmark to evaluate the quality of life journals. Experiment results show that \sysname can generate high-quality journals. We believe that life journaling represents a significant milestone application by integrating LLMs with sensor data, paving the way for new applications in personal daily life tracking and beyond. We will continue to enrich the dataset in future work.

\balance
\bibliographystyle{ACM-Reference-Format}
\bibliography{sample-base}




\end{document}